\documentclass[10pt,twocolumn,letterpaper]{article}

\usepackage{wacv}
\usepackage{times}
\usepackage{epsfig}
\usepackage{graphicx}
\usepackage{amsmath}
\usepackage{amssymb}
\usepackage{tabularx}

\usepackage{gensymb}
\usepackage{multirow}
\usepackage{booktabs}
\usepackage{pifont}
\usepackage{arydshln}
\usepackage{graphicx}
\usepackage{caption}
\usepackage{subcaption}
\usepackage{xspace}
\usepackage{xcolor}
\usepackage{dashbox}
\usepackage{enumitem}

\newcommand{\norm}[1]{\left\lVert#1\right\rVert}

\newcolumntype{Y}{>{\centering\arraybackslash}X}

\renewcommand{\subsubsection}[1]{\stepcounter{subsubsection}\vspace{5pt}\noindent\textbf{\thesubsubsection. \hspace{1mm} #1}\vspace{3pt}}

\usepackage[pagebackref=true,breaklinks=true,letterpaper=true,colorlinks,bookmarks=false]{hyperref}

\wacvfinalcopy 

\def\wacvPaperID{187} 

\ifwacvfinal\fi
\begin{document}

\title{Estimating 3D Camera Pose from 2D Pedestrian Trajectories}

\author{Yan Xu\\
Carnegie Mellon University\\
{\tt\small yxu2@andrew.cmu.edu}
\and
Vivek Roy\\
Carnegie Mellon University\\
{\tt\small vroy@andrew.cmu.edu}
\and
Kris Kitani\\
Carnegie Mellon University\\
{\tt\small kkitani@cs.cmu.edu}
}

\maketitle
\ifwacvfinal\thispagestyle{empty}\fi

\begin{abstract}
We consider the task of re-calibrating the 3D pose of a static surveillance camera, whose pose may change due to external forces, such as birds, wind, falling objects or earthquakes.  Conventionally, camera pose estimation can be solved with a PnP (Perspective-n-Point) method using 2D-to-3D feature correspondences, when 3D points are known.  However, 3D point annotations are not always available or practical to obtain in real-world applications.  We propose an alternative strategy for extracting 3D information to solve for camera pose by using pedestrian trajectories. We observe that 2D pedestrian trajectories indirectly contain useful 3D information that can be used for inferring camera pose.  To leverage this information, we propose a data-driven approach by training a neural network (NN) regressor to model a direct mapping from 2D pedestrian trajectories projected on the image plane to 3D camera pose. We demonstrate that our regressor trained only on synthetic data can be directly applied to real data, thus eliminating the need to label any real data. We evaluate our method across six different scenes from the \textit{Town Centre Street} and \textit{DUKEMTMC} datasets.  Our method achieves an improvement of $\sim50\%$ on both position and orientation prediction accuracy when compared to other SOTA methods.
\end{abstract}
\begin{figure}[t]
\begin{center}
\includegraphics[width=0.95\linewidth]{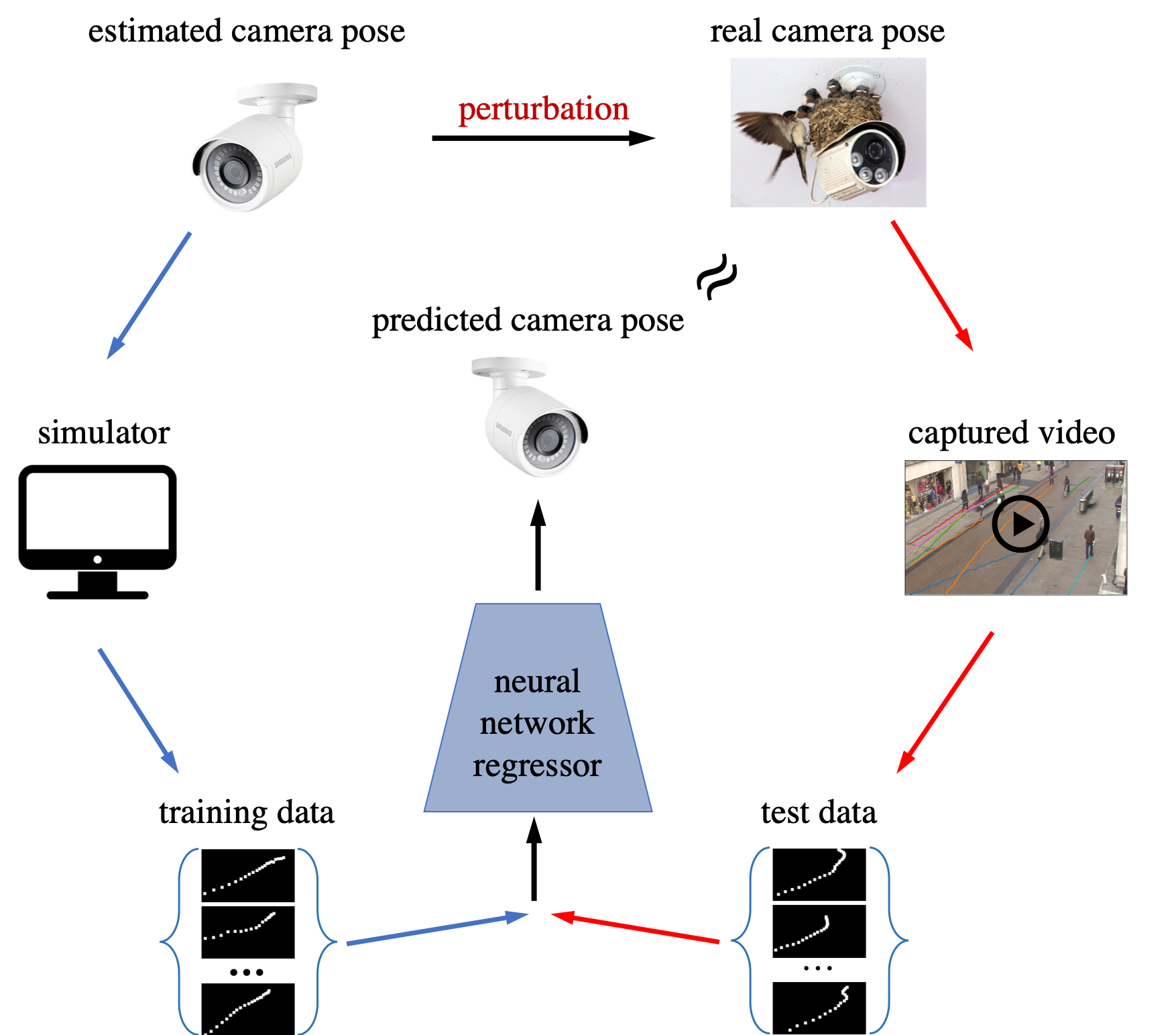}
\end{center}
  \caption{The structure of our proposed method. The training pipeline is connected with blue arrows, the test pipeline is connected with red arrows. The training data is synthetic data, the test data is extracted from real videos.\vspace{-5pt}}
\label{fig:tease_fig}
\end{figure}

\section{Introduction}
\label{sec:introduction}

Our task is to re-evaluate the 3D pose of a stationary single-view camera continually.  The camera pose changes over time because of external forces, including winds, earthquakes, and other factors.  For instance, a bird jumping on a calibrated security camera could cause a change in the looking angle of this camera.
Thus, the camera pose needs to be re-estimated.
PnP \cite{grunert1841pothenotische} methods seem to be a good solution to this problem except that they require 2D-to-3D correspondences.
Annotating 3D points is expensive, especially when continuous camera pose estimation is required.
Therefore in this task,  we assume that the 3D point annotations are unavailable and solve the problem with only 2D information inputs.
We find that 2D pedestrian trajectories contain useful information for estimating 3D camera pose.

According to the study from Carey \cite{carey2005establishing} on 815 pedestrians across ages, genders, fitness \etc, pedestrians walk in a roughly constant speed of about $1.4m/s$.
This statistical information can be leveraged for estimating camera pose.
If we assume that a pedestrian is walking at a strictly constant speed, the points on the 3D trajectory of this pedestrian should be equally spaced which can serve as a geometric constraint for solving camera pose.
Formally, we define the real camera pose as $\boldsymbol{\mathcal{P}^*}$, an estimated camera pose as $\boldsymbol{\tilde{\mathcal{P}}}$, and the pedestrian walking speed as $\boldsymbol{V}$.  We use $\boldsymbol{\tilde{\mathcal{P}}}$ to project the observed 2D pedestrian trajectory $\boldsymbol{\mathcal{T}}$ to the 3D ground.  As a result, when  $\boldsymbol{\tilde{\mathcal{P}}}$ is a close to $\boldsymbol{\mathcal{P}^*}$, points on the projected 3D trajectory should be equally spaced by $\boldsymbol{V}$.
Thus, we can build a mapping from a 2D pedestrian trajectory to the 3D camera pose: $\boldsymbol{\mathcal{P}^* = f(\mathcal{T}; V)}$.
However, the assumption that the pedestrian walks at a constant speed of $\boldsymbol{V}$ is too strong.
Instead, the walking speed of a pedestrian is unknown and unlikely to be constant over time.
Nevertheless, Carey \cite{carey2005establishing} shows that the statistical average of pedestrian speeds is generally constant ($\sim1.4m/s$) over time.  We thus propose a method to leverage the statistical information and solve the problem of camera pose estimation in a data-driven manner, by training a neural network regressor to learn an end-to-end mapping from a 2D trajectory to the 3D camera pose.

The structure of our proposed approach is presented in Figure \ref{fig:tease_fig}.
The input of our network is a 2D trajectory, consisting of points whose 2D coordinates represent the pedestrian position in the image.
The output of our network is the 3D camera pose, including position and orientation.
As shown in Figure \ref{fig:tease_fig}, our network is trained only with synthetic data.
For most data-driven methods, training data collection is costly.
We thus bypass this process by training our network with synthetic pedestrian trajectories.
Specifically, given a rough estimate of the camera pose and a human motion model, we can generate synthetic trajectories for training.
First, we sample a set of camera poses $\boldsymbol{\{P_i\}}$ around an estimation $\boldsymbol{P^{o}}$.  Our hypothesis is that the perturbation the real camera pose $\boldsymbol{P^*}$ is close to a camera pose in $\boldsymbol{\{P_i\}}$.
In the next step, we generate synthetic trajectories for each sampled camera pose $\boldsymbol{P_i}$ with the human motion model and use the data to train our NN regressor.
At test time, we extract real pedestrian trajectories from videos and feed them into the trained regressor.
We use the average of the predictions from all test trajectories as the estimated camera pose $\boldsymbol{\tilde{P}\approx P^*}$.
To verify the effectiveness, we evaluate our approach across six real camera settings from \textit{Town Centre Street} \cite{benfold2009guiding} and \textit{DUKEMTMC} \cite{ristani2016MTMC} datasets.  Compared to baseline methods using an image as input, our approach reduces the prediction error by $\sim50\%$ for both translation and rotation while exponentially decreasing training time cost. We summarize our contributions as follows.
\begin{enumerate}
\setlength\itemsep{0pt}
\vspace{-4pt}
\item[-] We propose an approach to regress the 3D pose of a stationary camera from only 2D pedestrian trajectories. Our approach could serve as a potential alternative solution for camera pose estimation, especially when 3D information is unavailable.
\vspace{-4pt}
\item[-] We demonstrate with experimental results that our NN regressor trained only on synthetic data can be directly applied to real data, therefore saving on the cost of collecting training data and has good generalizability.
\end{enumerate}

\section{Related Work}
\label{sec:related_work}

\noindent\textbf{Geometric Methods} are the conventional solution for camera pose estimation.  When 3D information is available, PnP methods \cite{grunert1841pothenotische} are the preferred choice.  These methods \cite{lepetit2009epnp, zheng2013revisiting, kneip2014upnp, campbell2017globally, liu2017efficient} solve for the camera pose by minimizing the reprojection error of 2D-to-3D feature correspondences, usually inside a RANSAC \cite{fischler1981random} loop to eliminate noisy data.  When 3D information is not available, methods have been proposed to employ constraints from geometric relationships, such as parallelism and orthogonality \cite{krahnstoever2005bayesian, wong2003camera,zhang2007camera, wong2010stratified, cao2006camera, colombo2006camera}, for estimating camera pose.  Geometric methods are usually accurate.  However, 2D-to-3D correspondences are not readily available in real applications, and geometric shapes are usually missing in many scenes.
To deal with this, we attempt to use a ubiquitous existing 2D information, pedestrian trajectories, to solve camera pose.

\begin{figure*}[t]
    \begin{subfigure}[b]{0.5\textwidth}
      \centering
        \includegraphics[width=\textwidth]{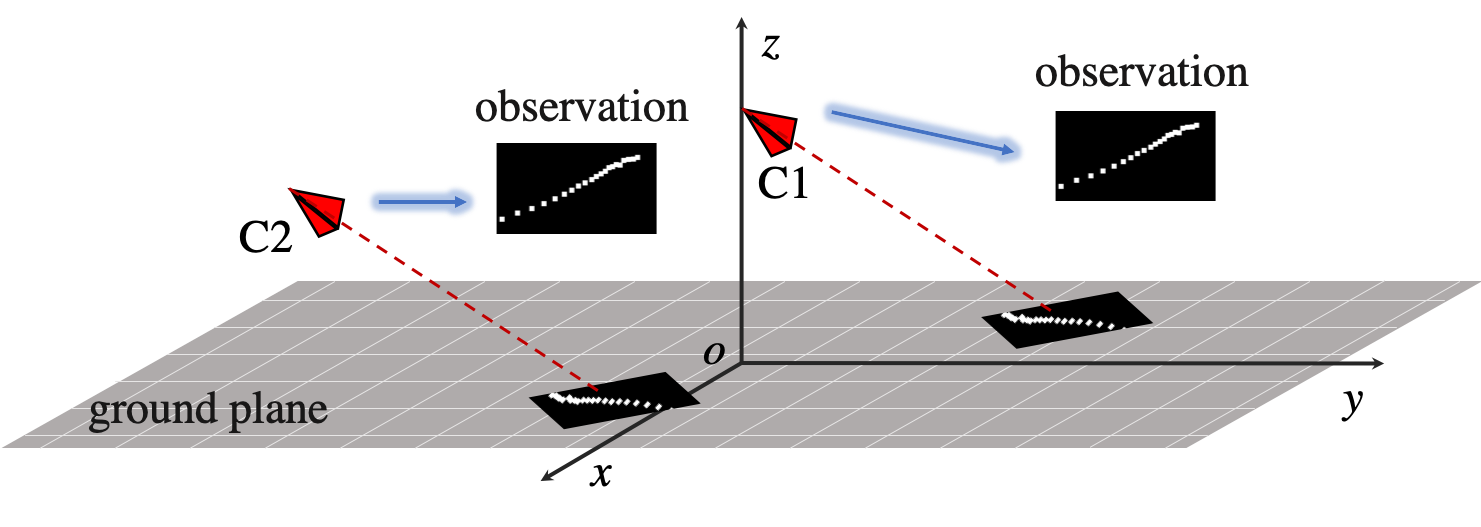}
        \captionsetup{width=\linewidth}
        \caption{Position ambiguity.}
        \label{subfig:pos_ambi}
    \end{subfigure}
    \begin{subfigure}[b]{0.5\textwidth}
      \centering
        \includegraphics[width=\textwidth]{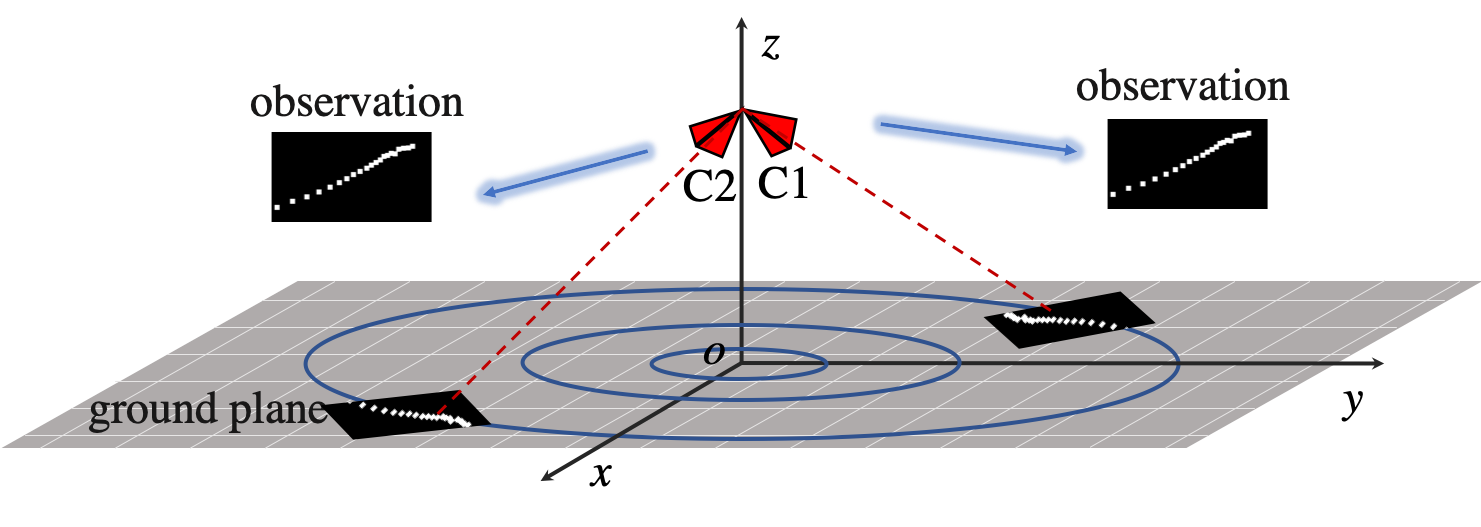}
        \captionsetup{width=\linewidth}
        \caption{Orientation ambiguity.}
        \label{subfig:rot_ambi}
    \end{subfigure}
\caption{Location and orientation ambiguities when estimating 3D camera pose with only pedestrian trajectories.  In each subfigure, \textit{C1} is the real camera pose, \textit{C2} represents an alternative fake camera pose that has the same observation as \textit{C1}. \vspace{-10pt}}
\label{fig:pred_ambi}
\end{figure*}

\vspace{1mm}
\noindent \textbf{Learning-based Methods} for camera pose estimation have shown encouraging results in recent years.
One class of learning-based methods regress camera pose end-to-end from an image or a video clip.
A representation of these methods is PoseNet \cite{kendall2015posenet}, which regresses the 6-DoF camera pose directly from a monocular image.
Since PoseNet, many methods have been proposed.  These methods either utilize different network architectures \cite{kendall2017geometric, wu2017delving, naseer2017deep, melekhov2017image} or leverage temporal information from the input \cite{walch2017image, clark2017vidloc} or introduce geometric constraints \cite{kendall2017geometric, laskar2017camera, brahmbhatt2018geometry}  to improve performance.
The second class of learning-based methods put a learnable neural network module into a structural pipeline, establishing correspondences between 2D pixels and 3D points, then solve camera pose by solving the RANSAC PnP problem.  These methods include DSAC/DSAC++ \cite{brachmann2017dsac, brachmann2018learning}, BTBRF \cite{meng2017backtracking}, InLoc \cite{taira2018inloc} \etc.
Some other works treat camera pose estimation as a multi-task learning \cite{radwan2018vlocnet++, lin2018deep} or a metric learning \cite{balntas2018relocnet} problem.
Generally, all these methods do not require annotated 2D-to-3D correspondences. However, training data for these methods is still expensive to collect.
To avoid such costs, Chen \etal \cite{chen2019sports} propose to use synthetic training images for Sports Camera Calibration.
We leverage a similar idea but use low-dimensional synthetic pedestrian trajectories as training data instead of high-dimensional images.

\vspace{1mm}
\noindent \textbf{Camera Pose Estimation Using Pedestrians} have been well investigated in the last decade.  Some methods \cite{rahimi2004simultaneous, anjum2011camera} simultaneously solve tracking and camera pose estimation using pedestrian trajectories by assuming a constant velocity of the moving object.
Other methods \cite{lv2002self, lv2006camera, krahnstoever2006autocalibration, junejo2006robust, liu2011surveillance} leverage the prior knowledge about the distribution of relative human heights to perform camera pose estimation.  A third class of methods \cite{guan2016extrinsic, tang2016camera, huang2016camera, brouwers2016automatic, lv2002self} make use of head and foot location information to calculate the vanishing points and line to further estimate camera pose.
Our work also leverages the geometric information in pedestrian trajectory to estimate the 3D camera pose. On contrary to these methods, we chose to solve the problem in a learning-based manner.  Our intuition is to leverage the good expressive ability of neural networks to learn a mapping from input to camera pose while improving noise-robustness.


\section{Approach}
\label{sec:approach}

As mentioned in Section \ref{sec:introduction}, the 3D pose of a static camera can be estimated from 2D pedestrian trajectories, under the assumption that the pedestrian is moving at a constant speed.
Formally, consider a pedestrian moving with a constant speed $\boldsymbol{V}$ on the ground plane $\boldsymbol{\pi_g}$ being observed by a camera.
We aim to build a mapping from a 2D projection of the trajectory $\boldsymbol{\mathcal{T}}$ to the 3D camera pose $\boldsymbol{\mathcal{P}}$,using a neural network approximation function $\boldsymbol{f(\cdot;\theta)}$, where $\boldsymbol{\theta}$ are learnable network parameters:
\begin{equation}
\boldsymbol{\mathcal{P} = f(\mathcal{T}(V);\theta)}
\label{eq:regress_function}
\end{equation}

In this section, we first make necessary explanations on the definition of camera pose $\boldsymbol{\mathcal{P}}$ and the evaluation metrics.  We then describe the proposed neural network architecture and give the loss which we use to supervise the learning.  Finally, we explain synthetic training data generation.

\subsection{Evaluation Metrics}

The 3D pose of a camera $\boldsymbol{C}$ is conventionally defined as the location and orientation of the camera in a world coordinate system.  The location of $\boldsymbol{C}$ is often specified by a vector $\boldsymbol{t\in\mathbb{R}^{3}}$ with respect to the world origin.  The orientation of $\boldsymbol{C}$ can be described with several representations, such as a rotation matrix, a set of Euler angles, or a quaternion.  Among these representations, the quaternion $\boldsymbol{q\in\mathbb{R}^{4}}$ is most commonly used to describe the camera orientation.  A tuple then represents the overall 3D camera pose:
\begin{equation}
\boldsymbol{\mathcal{P}=(t, q)}
\label{eq:cam_pose}
\end{equation}

Most deep pose regressors directly predict the camera pose $\boldsymbol{\mathcal{P}}$ as a 7-dimensional vector.  However, if the pedestrian trajectory is the only information input for estimating the camera pose, then there will be ambiguities in the predicted camera pose, as shown in Figure \ref{fig:pred_ambi}.   From Figure \ref{fig:pred_ambi}, we observe that the 2D trajectory can only decide 1-dimensional location (height) and 2-dimensional orientation (pitch and roll angle) of the camera.
In the implementation, we focus on the relative position between the camera and the scene and only predict the height of the camera $\boldsymbol{Z_c}$: $\boldsymbol{t = [Z_c]}$.  For the estimation of camera orientation, we still predict the quaternion $\boldsymbol{\mathbf{q}}$, but we fix the yaw angle at training and test.
(Note that: Other deep pose regressors also have the ambiguity problem.  However, since they use images as input and it is unlikely to get the same images at different places, so the ambiguity problem does not affect them much.  However, as a cost, lighting and seasonal changes will have dramatic impacts on those deep pose regressors.)

The performance of a pose estimator is estimated by the location prediction error $\boldsymbol{t_{err}}$ and orientation prediction error $\boldsymbol{r_{err}}$.  $\boldsymbol{t_{err}}$ is defined as the Euclidean distance between the real location $\boldsymbol{t^*}$ and the predicted location $\boldsymbol{t}$:
\begin{equation}
\boldsymbol{t_{err} = \norm{t^* - t}_2}
\label{eq:pos_err}
\end{equation}
$\boldsymbol{r_{err}}$ is measured by the angle between the real orientation $\boldsymbol{q^*}$ and predicted orientation $\boldsymbol{q}$ \cite{shavit2019introduction}:
\begin{equation}
\boldsymbol{r_{err} = 2\cos^{-1}(|\langle q^*, q\rangle|)}
\label{eq:ori_err}
\end{equation}

\subsection{Network Architecture}

\begin{figure}[t]
\begin{center}
\includegraphics[width=0.9\linewidth]{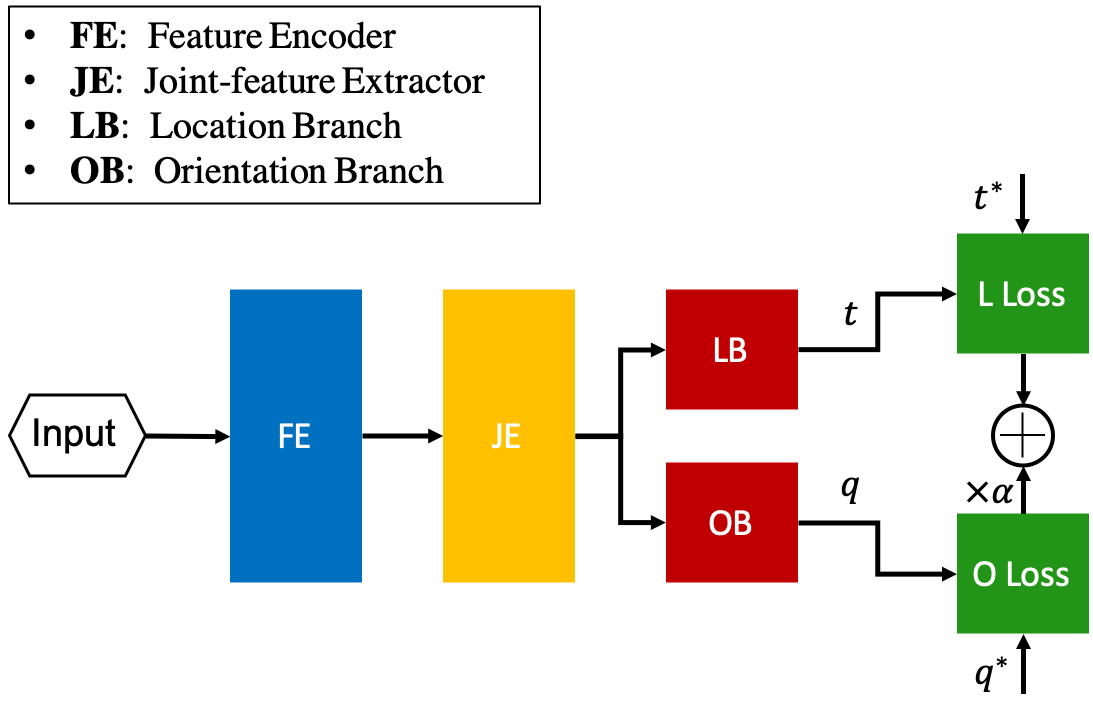}
\end{center}
  \caption{Architecture of our proposed NN regressor.}
\label{fig:network_structure}
\end{figure}

The architecture of our proposed NN regressor is presented in Figure \ref{fig:network_structure}.  The input of our network is a 2D pedestrian trajectory $\boldsymbol{\mathcal{T}\in\mathbb{R}^{2\times N}}$, where $\boldsymbol{N}$ is the length of the trajectory, and 2 is the dimension of the pixel coordinate of each point on the image plane.
The output of our network is the camera pose $\boldsymbol{\mathcal{P}}$ as given in Equation \ref{eq:cam_pose}.

Considering the input of our network is a trajectory with variable length, we first use an RNN Feature Encoder (FE) to encode the input trajectory $\boldsymbol{\mathcal{T}}$ to feature $\boldsymbol{u}$ with a fixed dimension.  After the FE module, we concatenate a Joint-feature Extractor (JE) to learn a common feature $\boldsymbol{v}$ that helps predict both location and orientation.  Finally, the camera location and orientation are separately predicted from $\boldsymbol{v}$ with a Location Branch (LB) and an Orientation Branch (OB).

\vspace{1mm}
\noindent\textbf{FE} is a bi-directional LSTM \cite{schuster1997bidirectional} with a $64$-dimensional hidden layer whose input is $\boldsymbol{\mathcal{T}\in\mathbb{R}^{2\times N}}$.  It has two hidden layers, and connects the output of the two hidden layers of opposite directions as a final output $\boldsymbol{u\in\mathbb{R}^{128}}$.  In the implementation, we found that bi-directional LSTM significantly outperforms single-directional LSTM in terms of the prediction accuracy of camera pose.

\vspace{1mm}
\noindent\textbf{JE} is a multi-layer perceptron (MLP) consisting of three fully connected layers.  The sizes of the three layers are $128\times256$, $256\times1024$, $1024\times512$ respectively.  The input is the feature $\boldsymbol{u\in\mathbb{R}^{128}}$ from FE, and the output is the joint-feature $\boldsymbol{v\in\mathbb{R}^{512}}$.

\vspace{1mm}
\noindent\textbf{LB and OB} are two branches taking $\boldsymbol{v}$ from JE as input.  LB is a 3-layer MLP with the sizes $512\times256$, $256\times128$, $128\times1$.  OB is another 3-layer MLP with the sizes $512\times256$, $256\times128$, $128\times4$.  The output of LB is camera location $\boldsymbol{t\in\mathbb{R}^{1}}$, and the output of OB is a quaternion $\boldsymbol{q\in\mathbb{R}^{4}}$ representing camera orientation.  We found that separately predicting $\boldsymbol{t}$ and $\boldsymbol{q}$ leads to higher prediction accuracy than predicting them together with one fully-connected layer.  Our intuition is that since $\boldsymbol{t}$ and $\boldsymbol{q}$ are measured with different scales (\textit{meter} vs \textit{radian}), it is better to predict them with separate branches.

All the fully-connected layers except for the last layers of LB and OB are concatenated with a non-linear layer.  The activation function is ReLU.

\subsection{Loss}

The regression function $\boldsymbol{f(\cdot;\theta)}$ from a trajectory $\boldsymbol{\mathcal{T}}$ to the camera pose $\boldsymbol{\mathcal{P}}$ is presented in Equation \ref{eq:regress_function}. We aim to solve for an optimal $\boldsymbol{\theta=\theta^*}$ such that the difference between the predicted camera pose $\boldsymbol{\mathcal{P}}$ and the real camera pose $\boldsymbol{\mathcal{P^*}}$ is minimized.  The loss function for supervising training is given in Equation \ref{eq:network_loss}. It consists of two parts: the Location Loss (L Loss) and the Orientation Loss (O Loss).
\begin{equation}
\boldsymbol{J(\mathcal{T}) = \underbrace{\norm{t^* - t}_2}_{Location Loss} + {\alpha}\cdot \underbrace{\norm{q* - \frac{q}{\norm{q}_2}}_2}_{Orientation Loss}}
\label{eq:network_loss}
\end{equation}
in which, $\boldsymbol{t^*}$ and $\boldsymbol{q^*}$ represent the real camera pose, and $\boldsymbol{\alpha}$ is an adjustable scale factor used to balance the prediction accuracies of location and orientation.  The first term describes the location loss, which is the Euclidean distance between the predicted location $\boldsymbol{t}$ and the real location $\boldsymbol{t^*}$.  The second term is the orientation loss, measuring the difference between the predicted quaternion $\boldsymbol{q}$ and the real quaternion $\boldsymbol{q^*}$. Euclidean distance is used to approximate the spherical distance for the orientation loss, which has been proved effective by Kendall \etal \cite{kendall2015posenet}.

The parameter $\boldsymbol{\alpha}$ in PoseNet needs to be carefully tuned for different scenes.  However, we found that the value of $\boldsymbol{\alpha}$ does not have much impact on the performance of our regressor.  We've trained networks across different camera pose settings using $\boldsymbol{\alpha\in[1, 10000]}$, and the difference in the prediction accuracies is negligible.  We guess it is because our input is a 2D trajectory, which contains less redundant information than a colorful image.  It is easier for the network to learn the difference between location and orientation, even without tuning $\boldsymbol{\alpha}$.  Therefore, we set $\boldsymbol{\alpha=1}$ in all our experiments.

\subsection{Synthetic Training Data}
\label{subsec:approach_train_test}
To save the cost of training data collection and annotation, we propose to train our NN regressor with synthetic data and directly apply to real data.  To generate synthetic training data, we consider the following scenario: 1) A non-accurate estimation of the camera pose is available. As an example, the estimation can be from past measurements.  2) The intrinsic parameters are known, which is reasonable since the intrinsic parameters will not change once the camera is produced.
In such a scenario, we can generate synthetic data for training.

\vspace{1mm}
\noindent\textbf{Training with Synthetic Data}  When a non-accurate estimation of the camera pose is available, we can hypothesize that the real camera pose is in a reasonably close range to the estimated one.  Let the estimated camera pose be $\boldsymbol{\mathcal{P}^o}$:
\begin{equation}
\begin{split}
&\boldsymbol{\mathcal{P}^o = (t^o, \theta^o)}\\
&\boldsymbol{t^o = (t^o_x, t^o_y, t^o_z)}\\
&\boldsymbol{\theta^o = (\theta^o_x, \theta^o_y, \theta^o_z)}
\end{split}
\end{equation}
in which $\boldsymbol{t^o}$ is the estimated location, and $\boldsymbol{\theta^o}$ is the estimated orientation.  We then sample a set of camera poses $\boldsymbol{\{\mathcal{P}_i\}}$ around $\boldsymbol{\mathcal{P}^o}$.  In implementation, we use the ground truth camera pose given in the dataset as $\boldsymbol{\mathcal{P}^o}$, around which we sample $\boldsymbol{\{\mathcal{P}_i\}}$ and make sure $\boldsymbol{\mathcal{P}^o}\notin\boldsymbol{\{\mathcal{P}_i\}}$.  In real application, $\boldsymbol{\mathcal{P}^o}$ can be obtained from a past measurement.

For sampling location, we keep $\boldsymbol{[t^o_x, t^o_y]}$ fixed and sample camera height from $\boldsymbol{[\min{(0, t^o_z-3m)}, t^o_z+3m]}$.  For sampling orientation, since we use the $x$-conventional Euler angles in the experiments, $\boldsymbol{\theta^o_x}$ represents the yaw angle.  We keep $\boldsymbol{\theta^o_x}$ fixed and sample other two angles from $\boldsymbol{[\tilde{\theta}-15^\circ, \tilde{\theta}+15^\circ]}$, in which $\boldsymbol{\tilde{\theta}\in\{\tilde{\theta}^o_y, \tilde{\theta}^o_z\}}$.
We make uniform samplings for both location and orientation and set the sampling steps to be $\boldsymbol{0.4m}$ and $\boldsymbol{2^\circ}$ respectively.

Once we have $\{\boldsymbol{\mathcal{P}_i}\}$, we can generate synthetic trajectories.  Considering that pedestrians generally walks with an constant average speed $\boldsymbol{V\approx 1.4m/s}$ \cite{carey2005establishing}, we assume $\boldsymbol{V}$ follows a Gaussian distribution:
\begin{equation}
\boldsymbol{V\sim\mathcal{N}(\bar{V}, \sigma^2)}
\end{equation}
in which $\boldsymbol{\bar{V}=1.4m/s$ and $\sigma=0.1m/s}$.  For each camera pose $\boldsymbol{\mathcal{P}_i}$, we randomly sample $10$ pedestrian speeds $\boldsymbol{\{V_j\}, j\in[1, 10], j\in\mathbb{R^N}}$.  By doing this, we can guarantee that $\boldsymbol{p(V_j\in[\bar{V}-3\sigma, \bar{V}+3\sigma])=99.74\%}$. Namely, the sampled pedestrian speeds will fall into the interval of $\boldsymbol{[1.1m/s, 1.7m/s]}$ with a probability of $\boldsymbol{99.74\%}$, which is reasonable in real-life scenarios.
For each combination of camera pose and pedestrian speed $\boldsymbol{\{\mathcal{P}_i, V_j\}}$, we generate 1 synthetic 2D trajectory $\boldsymbol{\mathcal{T}_{ij}}$ with the camera projection model and a human motion model:
\begin{equation}
\boldsymbol{\mathcal{T}_{ij} = g(\mathcal{P}_i, V_j)}
\end{equation}
in which $\boldsymbol{g(\cdot)}$ is the trajectory generation function, and $\boldsymbol{\mathcal{T}_{ij}\in\mathbb{R}^{2\times N}}$, $\boldsymbol{N\in[11, 31]}$ is the length of $\boldsymbol{\mathcal{T}_{ij}}$.

\begin{figure}[t]
    \begin{subfigure}[b]{\linewidth}
        \centering
        \includegraphics[width=\textwidth]{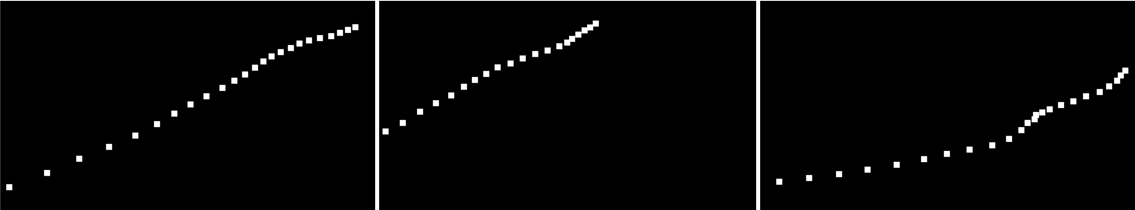}
        \captionsetup{width=\linewidth}
        \caption{Pedestrian trajectories from real video}
        \label{subfig:traj_real}
    \end{subfigure}

    \begin{subfigure}[b]{\linewidth}
        \centering
        \includegraphics[width=\textwidth]{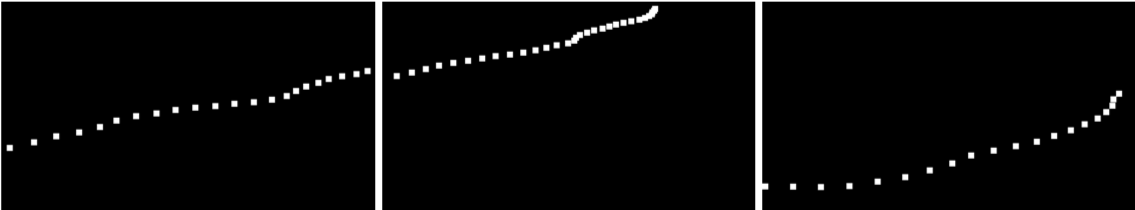}
        \captionsetup{width=\linewidth}
        \caption{Synthetic pedestrian trajectories}
        \label{subfig:traj_syn}
    \end{subfigure}
\caption{Pedestrian trajectories extracted from real videos and  synthetic trajectories generated from our simulator.\vspace{-5pt}}
\label{fig:syn_real_data}
\end{figure}

At training, we generate a synthetic dataset $\{\boldsymbol{\mathcal{T}_{ij}; \mathcal{P}_i}\}$ to train our NN regressor.
Figure \ref{fig:syn_real_data} presents a comparison between synthetic pedestrian trajectories and real pedestrian trajectories.  We can see that the synthetic data are almost indistinguishable from the real data.

\vspace{1mm}
\noindent\textbf{Test on Real Data}  At test time, we extract pedestrian trajectories $\{\boldsymbol{\mathcal{T}_{k}\}, k\in[1,K]}$, from videos captured by the camera, feed each $\boldsymbol{\mathcal{T}_{k}}$ to a trained regressor $\boldsymbol{f(\cdot;\theta^*)}$ to get a camera pose prediction $\boldsymbol{\tilde{\mathcal{P}}_{k}}$.  Finally, we leverage the prior knowledge on the statistical average of pedestrian speeds and take the mean of the predictions from all the trajectories as the final estimation of the real camera pose $\boldsymbol{\mathcal{P}^*}$:
\begin{equation}
\boldsymbol{\tilde{\mathcal{P}} = \frac{1}{K}\sum_{k=1}^K\tilde{\mathcal{P}}_{k} = \frac{1}{K}\sum_{k=1}^Kf(\mathcal{T}_{k}; \theta^*) \approx \mathcal{P}^*}
\end{equation}



\section{Experiments}
\label{sec:experiment}

In this section, we evaluate our approach on real camera settings and present the results and analysis.
Section \ref{subsec:real_dataset} introduces the datasets and necessary implementation details.
Section \ref{subsec:baseline_learning} compares our approach with learning-based baseline methods.
Section \ref{subsec:baseline_geometric} compares our approach with a geometric baseline method.
Section \ref{subsec:quantative_reprojection} gives a qualitative result of our approach.
Section \ref{subsec:ablation_study} provides the ablation study results.

\begin{figure}[t]
\includegraphics[width=\linewidth]{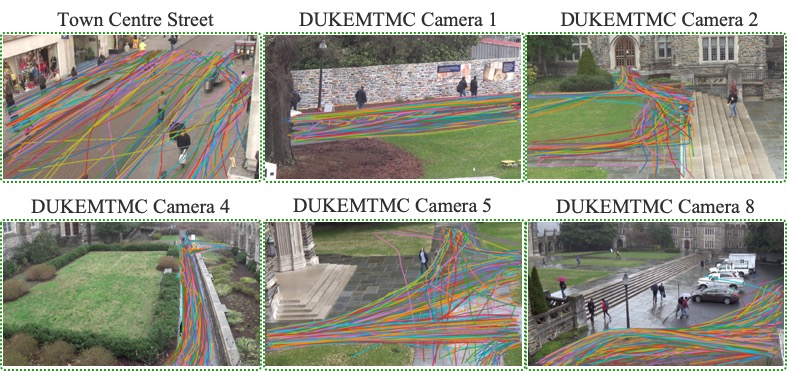}
\caption{Real pedestrian trajectories of for test. Only the trajectories on the same ground plane are used.\vspace{-5pt}}
\label{fig:real_scenes}
\end{figure}

\subsection{Datasets and Training Details}
\label{subsec:real_dataset}

The six real scenes that we used to evaluate our approach are from Town Centre Street \cite{benfold2009guiding} and DUKEMTMC \cite{ristani2014tracking}.  These two datasets contain a large number of pedestrians moving at roughly constant speeds.  Figure \ref{fig:real_scenes} shows the real pedestrian trajectories we use for testing.

\begin{table*}[h]
\footnotesize
\begin{center}
\renewcommand{\arraystretch}{1.3}
\begin{tabularx}{\linewidth}{|l l|Y Y Y Y Y Y|Y|}
\hline
\multicolumn{2}{|c@{\hspace{5mm}}|}{Method} & \shortstack{TCS} & \shortstack{DM1} & \shortstack{DM2} & \shortstack{DM4} & \shortstack{DM5} & \shortstack{DM8} & Average \\
\hline\morecmidrules\cline{1-8}
 & \multicolumn{1}{|r@{\hspace{5mm}}|}{$\alpha=1$} & $0.27m, 4.99^\circ$ & $0.74m, 5.08^\circ$ & $0.56m, 4.92^\circ$ & $0.26m, 5.26^\circ$ & $0.90m, 4.99^\circ$ & $0.35m, 4.45^\circ$ & $0.51m, 4.95^\circ$ \\
 & \multicolumn{1}{|r@{\hspace{5mm}}|}{10} & $0.23m, 5.13^\circ$ & \fbox{$0.42m, 4.89^\circ$} & $0.48m, 5.06^\circ$ & $0.22m, 5.11^\circ$ & $1.03m, 4.71^\circ$ & $0.24m, 4.97^\circ$ & $0.44m, 4.98^\circ$ \\
 & \multicolumn{1}{|r@{\hspace{5mm}}|}{50} & $0.29m, 4.96^\circ$ & $0.78m, 5.00^\circ$ & $0.57m, 5.03^\circ$ & $0.23m, 5.22^\circ$ & $0.87m, 4.57^\circ$ & $0.32m, 5.04^\circ$ & $0.51m, 4.82^\circ$ \\
 & \multicolumn{1}{|r@{\hspace{5mm}}|}{100} & $0.26m, 4.85^\circ$ & $0.43m, 4.99^\circ$ & $0.53m, 5.14^\circ$ & $0.25m, 5.06^\circ$ & $1.09m, 3.81^\circ$ & $0.29m, 5.05^\circ$ &$0.48m, 4.97^\circ$ \\
 & \multicolumn{1}{|r@{\hspace{5mm}}|}{500} & \fbox{$0.22m, 4.41^\circ$} & $0.51m, 5.01^\circ$ & $0.58m, 2.49^\circ$ & $0.19m, 4.99^\circ$ & $1.18m, 2.02^\circ$ & $0.27m, 4.81^\circ$ & $0.49m, 3.96^\circ$ \\
 & \multicolumn{1}{|r@{\hspace{5mm}}|}{1000} & $0.34m, 4.60^\circ$ & $0.63m, 4.68^\circ$ & $0.71m, 2.82^\circ$ & \fbox{$0.18m, 4.85^\circ$} & $0.92m, 1.87^\circ$ & $0.32m, 5.20^\circ$ & $0.52m, 4.00^\circ$ \\
 & \multicolumn{1}{|r@{\hspace{5mm}}|}{2000} & $0.26m, 4.62^\circ$ & $0.66m, 4.70^\circ$ & \fbox{$0.49m, 2.28^\circ$} & $0.33m, 3.76^\circ$ & $1.12m, 1.70^\circ$ & \fbox{$0.23m, 4.69^\circ$} & $0.52m, 3.63^\circ$ \\
\multirow{-8}{*}{{PoseNet~\cite{kendall2015posenet}}} & \multicolumn{1}{|r@{\hspace{5mm}}|}{5000} & $0.26m, 4.34^\circ$ & $0.96m, 5.10^\circ$ & $0.60m, 1.57^\circ$ & $0.29m, 3.46^\circ$ & \fbox{$0.86m, 1.86^\circ$} & $0.36m, 4.97^\circ$ & $0.56m, 3.55^\circ$  \\
\hdashline
\multicolumn{2}{|l@{\hspace{5mm}}|}{Best fine-tuned result} & $\textcolor{red}{0.22m, 4.41^\circ}$ & $\textcolor{red}{0.42m, 4.89^\circ}$ & $\textcolor{red}{0.49m, 2.28^\circ}$ & $\textcolor{red}{0.18m, 4.85^\circ}$ & $\textcolor{red}{0.86m, 1.86^\circ}$ & $\textcolor{red}{0.23m, 4.69^\circ}$ & $\textcolor{blue}{\mathbf{0.40m, 3.83^\circ}}$ \\
\hline\morecmidrules\cline{1-8}
\multicolumn{2}{|l@{\hspace{5mm}}|}{Ours} & $\mathbf{0.21m}, \mathbf{2.16^\circ}$ & $\mathbf{0.31m, 2.02^\circ}$ & $\mathbf{0.16m, 1.29^\circ}$ & $\mathbf{0.18m, 0.76^\circ}$ & $\mathbf{0.22m, 1.88^\circ}$ & $\mathbf{0.26m, 3.73^\circ}$ & $\textcolor{blue}{\mathbf{0.22m, 1.97^\circ}}$\\
\hline
\end{tabularx}
\end{center}
\caption{Location and orientation prediction errors of our approach and the PoseNet~\cite{kendall2015posenet} baselines with scale factor $\alpha$ of different values.  ``TCS'' denotes Town Center Street, ``DM$i$'' denotes DUKEMTMC scene-$i$. \textit{(The denotations will be the same for all following tables unless explicitly explained.)}   Our approach outperforms the best results of all baseline models.\vspace{-12pt}}
\label{tab:baseline_posenet}
\end{table*}

\vspace{1mm}
\noindent \textbf{Town Centre Street} (TCS) \cite{benfold2009guiding} is an outdoor dataset containing a video sequence of a busy street with up to thirty pedestrians visible at a time.  The video is 5 minutes long at 25 fps and 1080p resolution. Camera intrinsic and extrinsic parameters and bounding boxes for each pedestrian at each frame are available.

\vspace{1mm}
\noindent \textbf{DUKEMTMC} (DM) \cite{ristani2016MTMC} is a manually annotated and calibrated multi-camera dataset recorded outdoors on the Duke University campus with 8 synchronized cameras. It consists of 8 static cameras $\times$ 85 minutes of 1080p 60 fps video for a total of more than 2,000,000 manually annotated frames and more than 2,000 identities. The camera extrinsic and intrinsic parameters are provided. The ground plane coordinates of all the identities are accurately labeled.
We pick 5 of the most densely populated scenes from the total of 8 scenes to evaluate our method.\vspace{-2pt}

The model is implemented using PyTorch \cite{paszke2017automatic} and optimized using ADAM \cite{kingma2014adam} with $\boldsymbol{\beta_1=0.9$, $\beta_2=0.999}$, and an initial learning rate of $0.01$.  We train on one NVIDIA Titan X (Pascal) GPU with a batch size of 1024 and each round consisting of 50 epochs.

\subsection{Compare with Learning-based method}
\label{subsec:baseline_learning}

\begin{table}[t]
\footnotesize
\begin{center}
\renewcommand{\arraystretch}{1.3}
\begin{tabularx}{\linewidth}{|l| Y Y Y|}
\hline
Methods & TCS & DM1 & DM2 \\
\hline
DSAC++~\cite{brachmann2018learning} & $12.99m, 123.97^\circ$ & $23.57m, 99.55^\circ$ & $44.87m, 118.10^\circ$ \\
Ours & $\mathbf{0.21m}, \mathbf{2.16^\circ}$ & $\mathbf{0.31m, 2.02^\circ}$ & $\mathbf{0.16m, 1.29^\circ}$\\
\hline\morecmidrules\cline{2-4}
\multicolumn{1}{c|}{} & DM4 & DM5 & DM8 \\
\morecmidrules\cline{2-4}
\multicolumn{1}{c|}{} & $37.24m, 128.01^\circ$ & $35.46m, 104.08^\circ$ & $38.63m, 116.35^\circ$\\
\multicolumn{1}{c|}{} & $\mathbf{0.18m, 0.76^\circ}$ & $\mathbf{0.22m, 1.88^\circ}$ & $\mathbf{0.26m, 3.73^\circ}$\\
\morecmidrules\cline{2-4}
\end{tabularx}
\end{center}
\caption{Location and orientation prediction errors of DSAC++~\cite{brachmann2018learning} and our method on six real camera settings.  DSAC++ fails on this task, while our method performs well.\vspace{-5pt}}
\label{tab:dsac}
\end{table}

We compare our method with other learning-based pose estimators in this experiment. Current learning-based pose estimators can be generally divided into two classes. The first class of methods regress the camera pose end-to-end from an image or video clip, such as PoseNet \cite{kendall2015posenet} and its variants.
The second class of methods embed a learnable DNN module into a structural pipeline to solve camera pose, such as DSAC/DSAC++ \cite{brachmann2017dsac, brachmann2018learning}.
We use PoseNet and DSAC++ as the learning-based method baselines in this experiment.
Note that PoseNet and DSAC++ are used for solving the relocalization tasks, and their input data are images captured from different view angles.
However, in our task, the camera view angle does not change over time.
The observations at different time are identical except for the positions of the pedestrians.
To eliminate the impact of the static background on the prediction of the baseline methods, we set the input of the pose estimators to be a black and white image containing only a single pedestrian trajectory, as shown in Figure \ref{fig:syn_real_data}.

Table \ref{tab:baseline_posenet} gives a quantitative comparison of the predicted camera location error and orientation error between our approach and the PoseNet~\cite{kendall2015posenet} baselines with fine-tuning the scale factor $\boldsymbol{\alpha}$.  We observe that our proposed method almost \textit{doubles the prediction accuracy} on both location and orientation comparing to the best result of the fine-tuned baseline models.  Our model does not require fine-tuning $\boldsymbol{\alpha}$, saving the training cost.
Results show that 1) 3D camera pose can be estimated from only 2D pedestrian trajectories using learning-based methods and 2) our model trained on synthetic data can be directly applied to real test data.

Table \ref{tab:dsac} gives the quantitative comparison between DSAC++ \cite{brachmann2018learning} and our method.  We observe that SOTA DSAC++ fails on this task, while our method achieves low prediction errors.
We believe that DSAC++ fails because it relies on an image as input, which in our case is a mostly black image with sparse trajectory locations marked with white.
The first module of DSAC++, which is an FCN \cite{long2015fully}, maps from a 2D image to 3D points.
The black areas of different location inside an image or on the same locations of different images would have the same pixel values, zeros, but different depths.  However, the FCN is trained to map the pixels with the same zero values to 3D points with different depths, which leads to the collapse of DSAC++.

\begin{table}[t]
\small
\begin{center}
\renewcommand{\arraystretch}{1.3}
\begin{tabularx}{\linewidth}{|l| Y Y Y Y Y Y |}
\hline
Methods & TCS & DM1 & DM2 & DM4 & DM5 & DM8 \\
\hline
VP-1 & $86.22^\circ$ & $70.48^\circ$ & $46.11^\circ$ & $45.12^\circ$ & $32.81^\circ$ & $38.10^\circ$ \\
VP-2 & $84.28^\circ$ & $44.64^\circ$ & $55.37^\circ$ & $64.04^\circ$ & $28.26^\circ$ & $41.63^\circ$ \\
VP-3 & $89.22^\circ$ & $52.68^\circ$ & $38.37^\circ$ & $36.54^\circ$ & $19.65^\circ$ & $18.77^\circ$ \\
\hline
Ours & $\mathbf{2.16^\circ}$ & $\mathbf{2.02^\circ}$ & $\mathbf{1.29^\circ}$ & $\mathbf{0.76^\circ}$ & $\mathbf{1.88^\circ}$ & $\mathbf{3.73^\circ}$\\
\hline
\end{tabularx}
\end{center}
\caption{Orientation prediction errors of the vanishing-point calibration baseline~\cite{huang2016camera} and our approach.  Three different sets of orthogonal vanishing points are used, `VP-$i$' denotes the $i$-th set.  The output of baseline changes with different sets of VP. Our approach outputs stable accurate results.\vspace{-5pt}}
\label{tab:baseline_geometric}
\end{table}

\begin{figure*}[t]
\begin{center}
\includegraphics[width=\linewidth]{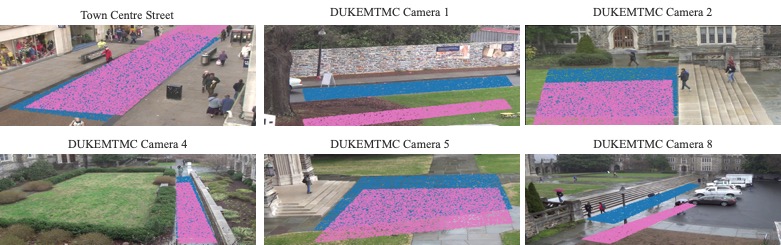}
\end{center}
  \caption{Visualization of the real ground plane and the re-projected ground plane.  The blue dotted plane in each image represents the real position of the ground plane, while the pink dotted plane represents the ground plane projected with the camera pose $\boldsymbol{\tilde{\mathcal{P}}}$ predicted by our NN regressor.  Our approach works reasonably well across all scenes.\vspace{-13pt}}
\label{fig:ground_projection}
\end{figure*}

\begin{figure}[t]
\includegraphics[width=\linewidth]{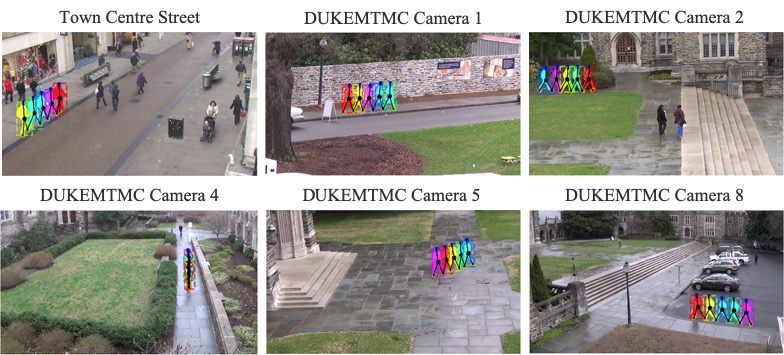}
\caption{The detected head and feet positions used in the vanishing-point calibration baseline.\vspace{-8pt}}
\label{fig:head_foot_detection}
\end{figure}

\subsection{Compare with Geometric method}
\label{subsec:baseline_geometric}

We compare our approach with traditional geometric methods in this section.  Geometric methods usually requires 2D-3D feature correspondences or objects with known geometric patterns, neither of which is available in our task.  However, Huang \etal~\cite{huang2016camera} proposed a method, using a pedestrian trajectory to find the three orthogonal vanishing points (VP) which can be applied for camera orientation (or rotation) estimation.  This method also leverage pedestrian trajectory information and can be applied in our experiment scenarios.  We compare our approach with the VP calibration method in this experiment.

Table \ref{tab:baseline_geometric} gives the rotation prediction errors of the VP calibration baseline~\cite{huang2016camera} and our approach. One set of orthogonal vanishing points can be detected from each walking pedestrian.  We use 3 different identities (or three different sets of orthogonal vanishing points) in this experiment.  We observe that the results computed by VP calibration baseline are inaccurate and variate with different sets of vanishing points, while our approach achieves accurate and stable results.  Our understanding is that, the VP baseline calculates a rotation between the camera coordinate system and the 3D coordinate defined by the three detected vanishing points.  With different sets of vanishing points, the results will be different.  On the contrary, our approach estimates the rotation between the camera coordinate system and the world coordinate system, and this rotation is unique and implicitly embedded in the training data.  Figure \ref{fig:head_foot_detection} shows the detected head and feet positions used in the VP baseline.

\vspace{-2pt}
\subsection{Qualitative Result}
\label{subsec:quantative_reprojection}

To obtain an intuitive understanding of the camera pose prediction error, we annotate the ground plane in each scene, project it to the 3D world with real camera pose  $\boldsymbol{\mathcal{P}^*}$, then re-project it back to the 2D image with our estimated camera pose $\boldsymbol{\tilde{\mathcal{P}}}$.
Figure \ref{fig:ground_projection} shows the result.  We observe that the re-projected ground planes generally overlap well with the real ground planes.  Note that, the re-projected ground plane not overlapping well with the ground truth does not means the prediction error is large, the visualization result also impact by the camera looking angle.  However, in general, our approach works reasonably well across all scenes.

\begin{figure}[t]
\begin{center}
\includegraphics[width=\linewidth]{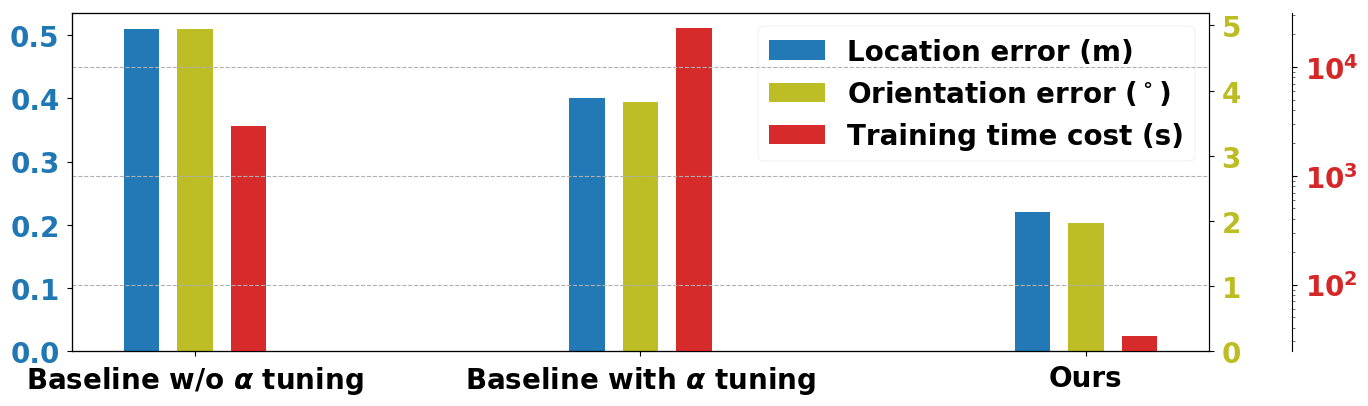}
\end{center}
  \caption{Efficiency comparison between our approach and baselines.  The left 3 bars are the result of baseline with $\alpha=1$, the middle 3 bars are the result of baseline with $\alpha$ fine-tuning, the right 3 bars are the result of our method.\vspace{-8pt}}
\label{fig:time_cost}
\end{figure}

\subsection{Ablation Study}
\label{subsec:ablation_study}
\vspace{-5pt}

\begin{figure*}[t]
\begin{center}
\includegraphics[width=\linewidth]{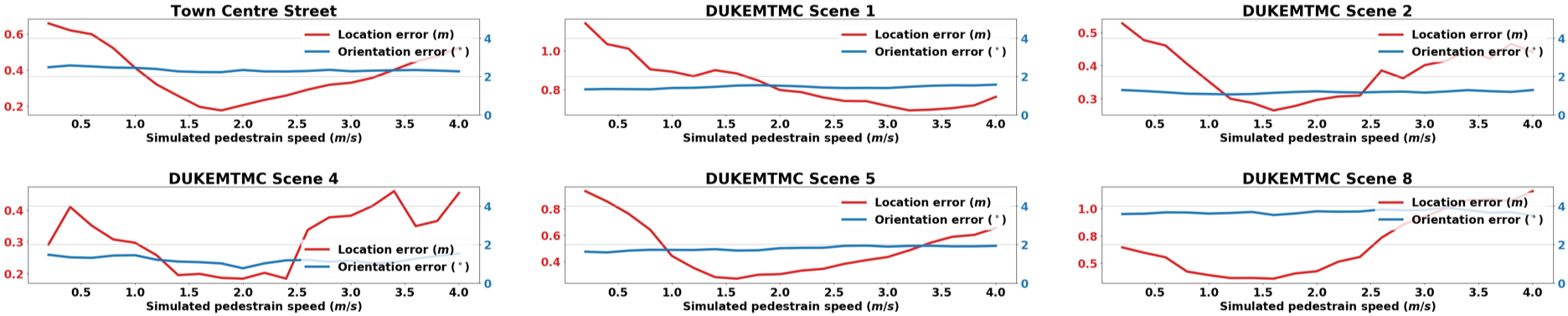}
\end{center}
  \caption{Location and orientation prediction errors of our approach over different synthetic speeds.  $x$-axis measures the synthetic speed, $y$-axis measures the prediction error. When the synthetic speed is close to human speed, the error is small.\vspace{-10pt}}
\label{fig:syn_speed}
\end{figure*}

\subsubsection{Analysis of Efficiency}
\label{sub:efficiency_analysis}

Figure \ref{fig:time_cost} shows the location and orientation prediction errors $vs$ the training time costs of our approach and the learning-based baselines.
For the baseline without fine-tuning $\alpha$, we set $\alpha=1$ as what we did for training our model.  We observe that, if we do not fine-tune $\alpha$, the baseline model will have high prediction errors on both location and orientation.  However, if we fine-tune $\alpha$, the training time cost will go up by almost an order.  Comparing with the baseline model with fine-tuning $\alpha$, our model reduces $\boldsymbol{50\%}$ of both the location and orientation prediction error.  Meanwhile, the training time cost of our model is exponentially less even than the baseline model without fine-tuning.

\subsubsection{Analysis of Trajectory Height Variation}

\begin{figure}[t]
\includegraphics[width=\linewidth]{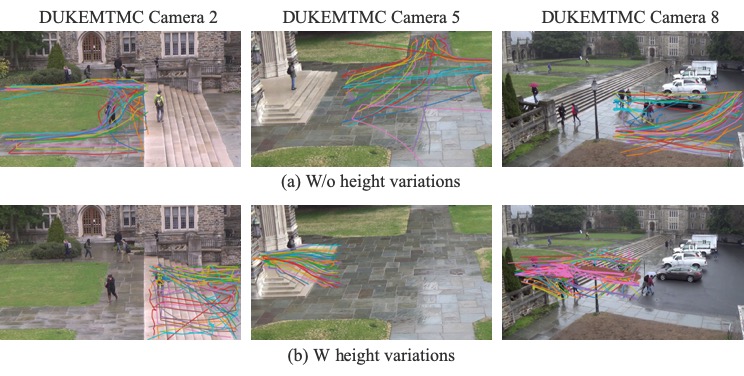}
\caption{Test trajectories with and without height variations used to test the same trained model.\vspace{0pt}}
\label{fig:traj_height}
\end{figure}

For a better understanding of how the height variations in the trajectories impact the performance, we test the same trained model on both trajectories that have height variations and trajectories that do not.  We conduct this experiment on DM2, DM5, and DM8, since the stairs in the three scenes will lead to significant trajectory height variations.
We generate two test sets for each scene and make sure the two sets have the same size.
One set contains trajectories passing through stairs, and another set contains trajectories on the ground plane.  The same trained model is tested on both sets.
A visualized comparison of the two test datasets is presented in Figure \ref{fig:traj_height}, and the result is given in Table \ref{tab:traj_height}.
We observe that height variation will damage the prediction accuracy, especially the location prediction accuracy.  Our explanation is, the stairs will impact the absolute distance of two adjacent points (step length) on the image plane. which will lead to high location (or depth) prediction error.

\subsection{Analysis of Synthetic Pedestrian Speed}
\label{subsec:exp_speed}

Since we use synthetic data to train our regressor, the distribution similarity between the synthetic trajectories and the real trajectories will, to a large extent, determine the performance of our approach.
To measure the impact of synthetic speed, we conduct the following experiment.
For each real test scene, we generate multiple training datasets, with the synthetic speed uniformly sampled from $\boldsymbol{[0.2m/s, 4.0m/s]}$.  The sampling step is $\boldsymbol{0.2m/s}$.  Then for each training dataset, we train a separate regressor and test the trained regressor with the same real test dataset.  We conduct this experiment on all the six scenes and present in Figure \ref{fig:syn_speed} the result.  We have the following observations:
\begin{itemize}
\setlength\itemsep{0pt}
\item[-] The synthetic speed has a significant impact on the location prediction accuracy while does not affect the orientation prediction accuracy that much.
\item[-] The location error plots for all the six datasets shape as a ``bowl'' over the synthetic speed.  We guess that the synthetic speed, which leads to the smallest location prediction error is the real average pedestrian speed. For 5 out of 6 datasets, the optimal synthetic speed is close to $1.4m/s$, which accords with Carey \cite{carey2005establishing}.
\item[-] The optimal synthetic speed for DM1 is abnormally around $3.5m/s$, which is almost as fast as the bicycle speed.  We guess that the given FPS parameter of DM1 is incorrect.  If we eliminate the result for DM1, the prediction errors will become $0.21m$ and $1.96^\circ$.
\end{itemize}

\begin{table}[t]
\small
\begin{center}
\renewcommand{\arraystretch}{1.3}
\begin{tabularx}{\linewidth}{|l| Y Y Y |}
\hline
Test Data & DM2 & DM5 & DM8 \\
\hline\morecmidrules\cline{1-4}
w/ & $0.22m, 1.31^\circ$ & $0.32m, 1.78^\circ$ & $0.38m, 3.49^\circ$ \\
w/o & $\mathbf{0.17m, 1.12^\circ}$ & $\mathbf{0.21m, 2.07^\circ}$ & $\mathbf{0.16m, 2.95^\circ}$ \\
\hline
\end{tabularx}
\end{center}
\caption{Position and orientation prediction errors when trajectories are w/ or w/o height variations.  The result shows trajectory height variations impact the performance.\vspace{0pt}}
\label{tab:traj_height}
\end{table}


\section{Conclusion}
\label{sec:conclusion}
In this paper, we have proposed a learning-based approach to end-to-end regress the 3D camera pose from 2D pedestrian trajectories.  Experiments on six real scenes has demonstrated that our approach can achieve high camera pose prediction accuracy across a variety of real-life camera settings.  We also verified with experiments that our proposed NN regressor could be trained on synthetic data only and directly applied to real test data.


{\small
\bibliographystyle{ieee}
\bibliography{egbib}
}

\newpage

\end{document}